\documentclass{article}

\usepackage{PRIMEarxiv}

\usepackage[utf8]{inputenc} 
\usepackage[T1]{fontenc}    
\usepackage{hyperref}       
\usepackage{url}            
\usepackage{booktabs}       
\usepackage{amsfonts}       
\usepackage{nicefrac}       
\usepackage{microtype}      
\usepackage{lipsum}
\usepackage{fancyhdr}       
\usepackage{graphicx}       
\graphicspath{{media/}}     

\usepackage[T1]{fontenc}
\usepackage{graphicx}
\usepackage{booktabs} 
\usepackage{amssymb}
\usepackage[acronym]{glossaries}
\usepackage{glossary-inline}
\usepackage{subcaption}
\usepackage{lscape} 
\usepackage{bigfoot}
\usepackage{blindtext}
\usepackage{makecell}
\usepackage[labelfont=bf]{caption}
\usepackage{setspace}
\usepackage{enumitem}
\makeglossaries
\newacronym{drl}{DRL}{Deep Reinforcement Learning}
\newacronym{rl}{RL}{Reinforcement Learning}
\newacronym{ppo}{PPO}{Proximal Policy Optimization}
\newacronym{re}{RE}{Renewable Energy}
\newacronym{ml}{ML}{Machine Learning}
\newacronym{ca}{CAgent}{CurriculumAgent}
\newacronym{pbt}{PBT}{Population Based Training}
\newacronym{cem}{CEM}{Category Embedding Model}
\newacronym{mdp}{MDP}{Markov Decision Process}
\newacronym{sgd}{SGD}{Stochastic Gradient Decent}
\newacronym{tso}{TSO}{Transmission System Operator}
\newacronym{opf}{OPF}{Optimal Power Flow}
\newacronym{l2rpn}{L2RPN}{Learning to Run a Power Network}
\newacronym{gnn}{GNN}{Graph Neural Network}
\newacronym{mst}{MST}{Median Survival Time}
\newacronym{tts}{TTs}{Target Topologies}
\newacronym{tt}{TT}{Target Topology}
\newacronym{mstcm}{MSTCM}{Median Survival Time across Chronic Medians}
\newacronym{sacd}{SACD}{Soft Actor-Critic Discrete }
\newacronym{senior}{$Senior_{95\%}$}{Senior Agent}
\newacronym{topo}{$TopoAgent_{85-95\%}$}{Topology Agent} %
\newacronym{senior85}{$Senior_{85\%}$}{85\%-Senior Agent}
\newacronym{d_n}{$DoNothing$}{Do-Nothing Agent}
\newacronym{smaac}{SMAAC}{Semi Markov Afterstate Actor Critic}
\newacronym{ddqn}{DDQN}{Dueling Deep Q Network}
\newacronym{bohb}{BOHB}{Bayesian Optimization Hyperband}
\newacronym{pca}{PCA}{Principal Component Analysis}
\newacronym{rf}{RF}{Random Forest}
\newacronym{xg}{XGBoost}{Extreme Gradient Boosting}
\newacronym{lgbm}{LightGBM}{Light Gradient-Boosting Machine} 
\newacronym{gandalf}{GANDALF}{Gated Adaptive Network for Deep Automated Learning of Features}
\newacronym{gflu}{GFLU}{Gated Feature Learning Units}
\newacronym{gru}{GRU}{Gated Recurrent Units}
\newacronym{mlp}{MLP}{Multi Layer Perceptron}
\newacronym{ood}{OOD}{out-of-distribution} 
\newacronym{tpe}{TPE}{Tree-structured Parzen Estimator} 
\newacronym{mcts}{MCTS}{Monte Carlo Tree Search}
\newacronym{fnn}{FNN}{Feed-forward Neural Network}
 
\newcommand*\samethanks[1][\value{footnote}]{\footnotemark[#1]}
  
\title{Fault Detection for agents on power grid topology optimization: A Comprehensive analysis
}

\author{
  Malte Lehna \thanks{Both authors contributed equally.\\ \texttt{malte.lehna@iee.fraunhofer.de}, \url{https://orcid.org/0000-0003-0621-1442} -\\ \texttt{mohamed.hassouna@iee.fraunhofer.de}, \url{https://orcid.org/0000-0001-9927-1625}} \\
  Fraunhofer IEE, Kassel University \\
  Kassel \\
  Germany\\
   \And
  Mohamed Hassouna\samethanks[1]\\
  Fraunhofer IEE, Kassel University \\
  Kassel \\
  Germany\\
  \And
  Dmitry Degtyar\\
  Fraunhofer IEE \\
  Kassel \\
  Germany\\
  \And
  Sven Tomforde\thanks{\url{https://orcid.org/0000-0002-5825-8915}} \\
  Kiel University: Intelligent Systems \\
  Kiel \\
  Germany\\
    \And
  Christoph Scholz\thanks{\url{https://orcid.org/0000-0002-8719-8261}} \\
  Fraunhofer IEE, Kassel University \\
  Kassel \\
  Germany\\
}

\begin{document}
\maketitle

\begin{abstract}
Optimizing the topology of transmission networks using \gls{drl} has increasingly come into focus. Various \gls{drl} agents have been proposed, which are mostly benchmarked on the Grid2Op environment from the \gls{l2rpn} challenges. The environments have many advantages with their realistic grid scenarios and underlying power flow backends. However, the interpretation of agent survival or failure is not always clear, as there are a variety of potential causes. In this work, we focus on the failures of the power grid simulation to identify patterns and detect them in advance. We collect the failed scenarios of three different agents on the WCCI 2022 \gls{l2rpn} environment, totaling about 40k data points. By clustering, we are able to detect five distinct clusters, identifying common failure types. Further, we propose a multi-class prediction approach to detect failures beforehand and evaluate five different prediction models. Here, the \gls{lgbm} shows the best failure prediction performance, with an accuracy of 82\%. It also accurately classifies whether a the grid survives or fails in 87\% of cases. Finally, we provide a detailed feature importance analysis that identifies critical features and regions in the grid.

\end{abstract}

\section{Introduction}\label{sec:intro}
With the increase of renewables in the energy mix and the resulting variability, the optimization of power grids has become increasingly complex. One possible approach is topology optimization, i.e., switching buses at a substation level, which is a more cost-effective alternative to redispatching \cite{marot2020learning} and can improve grid stability \cite{bacher1986network}. However, identifying the correct topology options is computationally expensive \cite{marot2020learning}. One possible solution for the topology optimization could be \glsfirst{drl}, as it can grasp large states and cope with the complexity \cite{viebahn2022potential}. 

Within the \gls{ml} research community, the \gls{l2rpn} challenges are the preferred research benchmarks \cite{kelly2020reinforcement,marot2022learning,marot2021learning,marot2020learning,9494879}. The underlying Grid2Op package of the \gls{tso} RTE provides several environments for \gls{drl} approaches.\footnote{Grid2Op: \url{https://grid2op.readthedocs.io/en/latest/} (last accessed 09/06/2024).} 
Within the environments, researchers train their agents on various chronics, which are scenarios containing one week of power grid data, including generator injections, load consumption, and structural information like planned outages and unexpected hazards, to identify effective policies. To rate agents, a scoring function is provided \cite{marot2021learning}, which depends on the agent's ability to maintain grid stability for a week in the test chronics. Several topology-based \gls{drl} agents have been applied to the \gls{l2rpn} challenges of RTE in the last years \cite{marot2020learning}. However, it is not always clear why an agent survives or fails in a specific chronic, as the environments contain stochastic components that change the behavior on repeated runs. Therefore, we analyze different grid failures in the IEEE118 grid and provide a predictive approach to detect possible failures in advance. 
\section{Contribution}\label{sec:contribution}
We take a closer look at agent failures to address the types of failures that can occur when using a topology-based agent. To do so, we aggregate the data of over 40k failed chronics from three different agents and cluster the failure types to identify specific patterns. 

Next to clustering, we provide a multi-class prediction framework to identify whether an agent is currently in a situation where survival can be ensured, or if failure is imminent. Based on the observation of the grid state, we can detect whether the grid will survive or if a failure will occur in 5, 3, or 1 time-steps, corresponding to 25, 15, and 5 minutes respectively. We test different prediction models and are able to identify with 87\% binary accuracy, whether grid failure is imminent. With a detailed feature importance analysis, we are further able to detect critical features and components in the electricity grid that are crucial for failure prediction.
Overall, the contributions can be summarized as follows: 
\begin{itemize}
    \item We evaluate three agents from Ref. \cite{lehna2024hugo} with ten different seeds on the IEEE118 grid (WCCI \gls{l2rpn} training environment). With 1662 chronics, we create a detailed dataset of around \textbf{40k grid failures}.
    \item We perform a \textbf{clustering analysis} that identifies distinct grid failure types and categorize them into five meaningful clusters, enhancing understanding of agents' shortcomings and their association with specific failure types. 
    \item We build a \textbf{multi-class forecasting framework} to identify the risk of failure beforehand. 
    \item We test five different forecasting models and identify the \textbf{\glsfirst{lgbm} as the best performing candidate}.
    \item With an additional \textbf{feature importance} analysis, we are able to identify critical features and regions within the IEEE118 grid. 
\end{itemize}
The rest of the paper is structured as follows. We provide an overview of the related work in Section \ref{sec:rel_work} and our methodology in Section \ref{sec:method}. We then present the experimental setup in Section \ref{sec:experiment}, followed by the results in Section \ref{sec:res}. Finally, we discuss our results in Section \ref{sec:discussion} and conclude in Section \ref{sec:conclusion}.

\section{Related Work}\label{sec:rel_work}
With the \gls{l2rpn} challenges, several approaches have been proposed to optimize the topology of transmission grids with \gls{drl}. In most cases, the \gls{drl} algorithms are combined with heuristic components \cite{chauhan2022powrl,binbinchen,lehna2024hugo,lehna2023managing} to increase performance and reduce the action space. Other researchers prefer to use evolutionary algorithms with planning components \cite{zhou2021action} or \gls{mcts} \cite{dorfer2022power}. Additionally, some approaches use \glspl{gnn} to represent the graph-structured state information \cite{grlsurvey2024}. However, most researchers report only the score and in some cases the survival of their agent on the test chronics. Understandably, they often do not show the shortcomings of their agents, making comparisons or even fault interpretation difficult. 
The most detailed analysis of the environments and agents comes from the \gls{l2rpn} challenges themselves, e.g., in \cite{marot2021learning,marot2020l2rpn}. They analyze the behavior of different agents and also show some examples using the Grid2Viz package.\footnote{Grid2Viz \url{https://github.com/rte-france/grid2viz} (last accessed 09/06/2024).} Alternatively, \cite{amdouni2023grid2onto} introduce Grid2Onto for Grid2Op environments. Their application is a recommender system based on a semantic model to display and select propositions from an agent for the respective grids. 

With respect to fault prediction, previous studies have investigated electricity grids, as evidenced by the surveys for transmission grids \cite{chen2016fault} and distribution grids \cite{dashti2021survey}. However, to the best of our knowledge, there is no specific work on fault prediction using \gls{drl} agents for topology optimization. The closest work was done by \cite{mu2022cascading}, who used an LSTM model to predict failures on the IEEE 35 bus network. However, they focus more on interlocking and cascading failures and not directly on the outcome of \gls{drl} agents. 
Therefore, there is still a need for further research in this area. 
\section{Methodology}\label{sec:method}
\subsection{Descriptive Analysis with Clustering}\label{ssec:cluster}
In the descriptive analysis, we want to identify different types of grid failures to better understand the shortcomings of the agents. However, the underlying data is high-dimensional and requires reduction. 
To ensure that we have less correlation in our clustering, we use the \gls{pca} \cite{abdi2010principal} with standardized variables. The dimensionality is reduced by transforming the data onto the coordinate system of the principal components, which are orthogonal to each other. 
We then select all major components that added at least 5\% of the variance to the data, thus reducing the dimensions effectively. In our case, this yields six components with an explained variance of 85\%.\\
With a reduced number of components, we combine the \gls{pca} with k-means clustering, similar to \cite{ding2004k}. Introduced by \cite{macqueen1967some}, k-means clustering divides the data into k-clusters, where each data point is assigned to the nearest cluster centroid based on its distance. By iteratively assigning precise centroids, the distances within the cluster is minimized. Since the cluster size k has to be chosen, we iterate over different sizes and choose the cluster size based on the inertia and the silhouette metric \cite{kodinariya2013review}. 

\subsection{Forecasting Models}\label{ssec:models}
In our forecasting framework, we go beyond a binary prediction of failure versus non-failure of the grid. Instead, we adopt a multi-class approach, forecasting potential failures at three distinct time steps before they occur, namely five, three and one time-steps ahead. This allows for a more granular understanding of the impending failures and provide insights into the decision-making of the models such as the average probability distribution among all classes. We employ the following five prediction models and compare their performance. In particular, we utilize gradient boosting techniques for their strong performance across various prediction tasks and their ability to easily provide insights into feature importance.

\begin{description}
    \item [\gls{rf} \cite{rf}] is an ensemble learning method that constructs simple decision trees during training and outputs the majority vote of their predictions for classification tasks. This approach enhances model accuracy and robustness by reducing overfitting and improving generalization compared to individual decision trees. 
    
    \item [\gls{xg} \cite{xgboost}] uses gradient boosting to iteratively build decision trees, such that each new tree is trained to correct the residual errors of all previous trees. It employs advanced techniques like regularization, weighted quantile sketch, and sparsity awareness to improve model accuracy, prevent overfitting, and handle large datasets efficiently. 
    
    \item [\gls{lgbm} \cite{lgbm}] is an ensemble learning method for classification that builds decision trees using a gradient boosting framework, optimizing for speed and memory efficiency. Unlike \gls{xg}, \gls{lgbm} uses a histogram-based algorithm and leaf-wise tree growth, which significantly enhances training speed and reduces memory usage, especially for large datasets. 
    
    \item [\gls{cem} \cite{pytorchtabular}] is a basic yet effective feed-forward neural network. Its architecture integrates categorical features through a learnable embedding layer, making it a suitable starting point and baseline for comparison with other models.
    
    \item [GANDALF \cite{gandalf}] stands for Gated Adaptive Network for Deep Automated Learning of Features and optimizes feature selection and engineering using \gls{gflu}. \gls{gflu} leverages a gating mechanism 
    to iteratively refine feature representations and filter out noisy data, as well as utilizes stage-specific feature masks, allowing for hierarchical and adaptive feature selection. This results in a robust feature representation, which is then used by an \gls{fnn} to produce final predictions \cite{gandalf}.

\end{description}
For the hyperparameter tuning of all models, Optuna \cite{optuna} is employed to facilitate an efficient hyperparameter search. We utilize \gls{tpe} \cite{tpe} as our optimization method, as it maintains two probability distributions: one for good configurations and another for all sampled configurations. By focusing on promising regions of the hyperparameter space, \gls{tpe} directs the search more efficiently than traditional methods like grid or random search. 

\section{Experimental Setting}\label{sec:experiment}
\subsection{Agents}\label{ssec:agents}
In order to provide a comprehensive analysis of agent failures, it is essential to look at different agents. While there are numerous agents available for consideration, we decided to use the agents of \cite{lehna2024hugo} since their data was already available to us. The three agents in question are: 
\begin{description}
    \item[\gls{d_n}] The first agent does not interfere with the grid and only provides DoNothing actions. This agent is considered the baseline and we expect a lot more failures from it. 
    \item[\gls{senior}] This agent is based on the CurriculumAgent \cite{lehna2023managing}, which selects topology actions on individual substations, when the threshold of $\rho_{max,t} >=0.95$ is breached. The agent further includes rule-based components, such as periodic reversion to the starting topology \cite{lehna2024hugo,lehna2023managing} and line reconnection. The agent is more advanced, but changes the substations quite often in uncertain events.
    \item[\gls{topo}] The last agent is an extension of the \gls{senior} and is described in detail in \cite{lehna2024hugo}. The main difference is that the agent searches for overall suitable \gls{tts} when the $\rho_{max,t}$ is in the interval of $0.85<\rho_{topo}<0.95$. In \cite{lehna2024hugo}, the \gls{topo} was able to achieve better results, thus we expect differences to the \gls{senior}.
\end{description}

\subsection{General Data Collection}\label{ssec:grid2op}
To ensure a in-depth analysis, it is useful to look at a wide range of different environment scenarios. For this reason, we chose the WCCI 2022 \gls{l2rpn} environment \cite{serre2022reinforcement}, with a total of 1662 chronics. 
These chronics are created with an expected electricity mix of 2050, i.e. with only 3\% fossil fuels remaining in the electricity mix \cite{serre2022reinforcement}, so more variability is expected. 
Given the stochastic components of the environment, we run each chronic a total of 10 times with different master seeds for each agent.\footnote{To ensure that there was no cherry picking, we chose the ten seeds completely random using \verb|np.seed|($8888$). For each seed, we also calculated the \verb|_statistics_l2rpn_dn| and \verb|_statistics_l2rpn_no_overflow_reco$| separately.} Afterwards, the failed chronics were selected. 
\paragraph{Cluster Data:} For the clustering approach, we want to compare the observation of the failure with the initial situation of the environment. Thus, we collected the observation at the start of each chronic $obs_{t=0}$ and one time step before failure $obs_{t=n-1}$. Since each observation has $4295$ variables, we aggregate the variables. First, since line capacity $\rho$ is an important indicator of grid stability, we collect the maximum and average line capacity $\rho_{max}$ and $\rho_{mean}$ of the $obs_{t=n-1}$. Additionally, we counted the number of disconnected lines $\#lines_{dis}$ and the total $ts_{overflow}$, representing the cumulative duration during which the lines $l \in L$ have been in an overflown state ($\rho_l \geq 1.0$). Second, we account for the number of substation changes $\#sub_{changed}$, by comparing the switched substations to $obs_{t=0}$. 
Third, it is necessary to account for the different power flow changes in the grid. For the load consumption, the generator injections, and both sides of the lines (origin and extremity), we record the maximal active load value $(load^p_{max},gen^p_{max},line^{ex,p}_{max},line^{or,p}_{max})$ and the average value $(load^p_{mean},gen^p_{mean},line^{ex,p}_{mean},line^{or,p}_{mean})$ as a fraction to the first observation $obs_{t=0}$.\footnote{We excluded reactive power flow and voltage angle due to high correlation with active power flow or constant values. Elements with a zero in $obs_{t=0}$ were also excluded to avoid mathematical errors. See appendix table \ref{tab:cluster_variables} for a summary of all variables.} The fraction is calculated for the loads, generators, and lines. As an example, we show the calculation for the $load^p_{max}$ and $load^p_{mean}$ with $\gamma$ being one load in the set of all loads $\gamma \in \Gamma$:

\begin{align}
    load^p_{max} &= \max_{\gamma \in \Gamma}\left[load^p_{\gamma,t=n-1}/load^p_{\gamma,t=0}\right] -1 \\ 
    load^p_{mean} &= \frac{1}{|\Gamma|} \sum^\Gamma_{\gamma=0} (load^p_{\gamma,t=n-1}/load^p_{\gamma,t=0} ) -1 
\end{align} 
Next to the cluster variables, we also collected the survival times ($t_{survived}$) as well as other descriptive variables for our analysis. 
\paragraph{Forecasting Data:} For the forecast, we distinguish between failure and survival scenarios. For failures, we focus on observations at time steps $t=n-1$, $t=n-3$, and $t=n-5$, with $n$ being the failed time step. 
For survived observations, we sample from the chronics observations with a high $\rho_{max,t}$. However, to avoid overlapping, the $obs_{t=survived}$ must be at least $n-6$ away from the failed time step. To avoid data leakage, we split the chronics beforehand into train, validation, and test sets, ensuring that no earlier data point from the test or validation set is found in the training set.
This approach yielded $189\,563$ observations, with half of those surviving ($obs_{t=survived}$), and the remaining half is split evenly among $obs_{t=n-1}$, $obs_{t=n-3}$, and $obs_{t=n-5}$. Subsequently, through the sampling different chronics, the data was divided into a training set of size $152\,234$, a validation set of size $18\,881$, and a testing set of size $18\,448$. Thanks to the courtesy of RTE, the validation environment of the WCCI 2022 challenge was made available to us. Consequently, our findings can also be validated on an \gls{ood} sample of size $6036$.
\subsection{Metrics}\label{ssec:metric}
We evaluate the performance of our prediction models using several key metrics. First, we include the \textbf{Accuracy}, which is the ratio of correctly predicted instances to the total instances. This is the metric on which all models are trained. Further, with the multi-class classification problem, we also include the \textbf{Balanced Accuracy} to address the imbalance in the data sets. It is defined as the average of recall obtained for each class. The \textbf{F1 score}, which can be interpreted as a harmonic mean of precision and recall, reaches its best value at 1 and worst score at 0. 
For our specific use-case, we also want to measure the \textbf{Binary Accuracy}. This metric is defined by the accuracy of the one-vs-all model that categorizes all failures into a single class and survival into another. It effectively assessing the model's ability to distinguish between survival and failure observations. 


\section{Results}\label{sec:res}
\subsection{Clustering Results}\label{ssec:cl_res}
\begin{figure}[!h]
    \centering
    \includegraphics[trim={0.8cm 1.5cm 0.0cm 0cm},clip, width=1.0\linewidth]{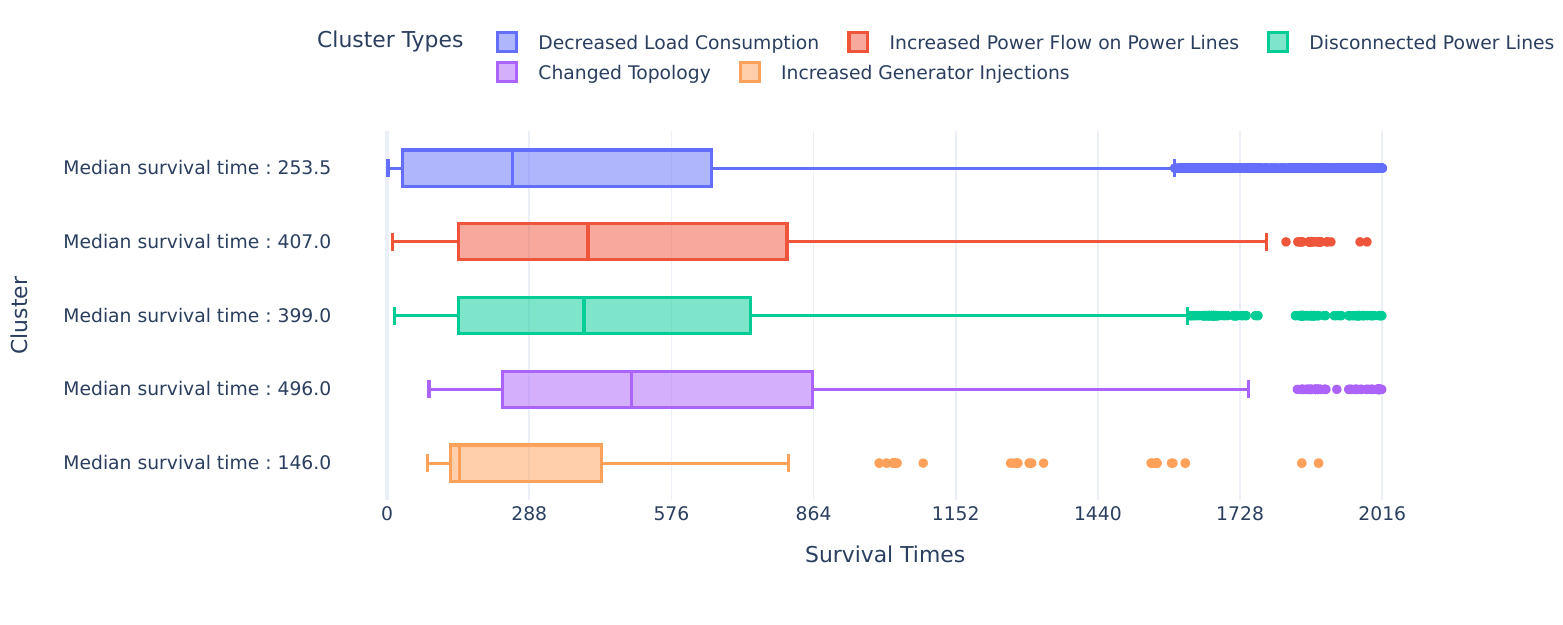}
    \caption{\small Box plot of the survival time $t_{survived}$ of each cluster. We further report the median survival time. The ticks on the x-axis correspond to a full day, e.g., 288 steps. Note that an agent must survive a total of 2016 time steps to complete the chronics successfully.}
    \label{fig:survival}
\end{figure}

\paragraph{Cluster Identification:} With the aggregated clustering data, we perform \gls{pca} and select six principal components, describing 85\% of the variance. Afterwards, we execute the k-means clustering and identify five clusters as the best cluster size for our data.\footnote{The plot of both inertia and silhouette can be found in the appendix figure \ref{fig:intertia}.} 
Subsequently, the clusters are grouped and the statistical components, including the mean, median, and quantiles of the variables, are analyzed to identify the distinctive characteristics of the clusters. We provide in Table \ref{tab:cluster_res} the mean values of all cluster variables.
The following clusters were identified:
\begin{enumerate}
    \item \textbf{\textit{Changed Topology}}: This cluster has the highest $\#sub_{changed}$ among all clusters.
    \item \textbf{\textit{Decreased Load Consumption}}: In this cluster the distinct characteristic is the lower load consumption of $load^p_{mean}$ and $load^p_{max}$.
    \item \textbf{\textit{Disconnected Power Lines}}: With the highest $ts_{overflow}$, $\#lines_{dis}$ and $\rho_{max}$ this cluster is directly linked to the overload of the power lines. 
    \item \textbf{\textit{Increased Generator Injections}}: This cluster has by far the highest generator values of $gen^p_{mean}$ and $gen^p_{max}$. 
    \item \textbf{\textit{Increased Power Flow on Power Lines}}: As the last cluster we identify the highest power flow in $line^{ex,p}_{max},line^{or,p}_{max},line^{ex,p}_{mean}$ and $line^{or,p}_{mean}$ as distinct feature. 
\end{enumerate}
\begin{table}[h]
    \centering
    \small
    \caption{\small Mean values of the clustering variables, grouped by one of the five clusters. Based on multiple statistical characteristics, we name the clusters accordingly. We highlight the highest mean value of each variable.}
    \begin{tabular}{lccccc}
        \toprule
        Cluster &  \makecell{Changed \\Topology}    & \makecell{Decreased\\ Load\\Consumption} & \makecell{Disconnected\\Power Lines} & \makecell{Increased\\ Generator\\Injections} & \makecell{Increased \\Power Flow\\on Power Lines}\\
        \midrule
        $ts_{overflow}$ & 5.32 & 6.89 & \textbf{12.60} & 10.32 & 9.96 \\
        $\#lines_{dis}$ & 0.79 & 1.58 & \textbf{3.93} & 1.37 & 2.12 \\
        $\#sub_{changed}$ & \textbf{5.54} & 1.71 & 0.92 & 2.94 & 3.57 \\
        $\rho_{max}$ & 1.03 & 1.13 & \textbf{1.43} & 1.25 & 1.24 \\
        $\rho_{mean}$ & 0.40 & 0.38 & 0.36 & \textbf{0.41} & 0.38 \\
        $load^p_{max}$ & 0.33 & 0.09 & \textbf{0.35} & 0.34 & 0.34 \\
        $load^p_{mean}$ & 0.12 & -0.06 & 0.13 & \textbf{0.14} & 0.13 \\
        $gen^p_{max}$ & 4.78 & 2.69 & 3.66 & \textbf{614.55} & 2.31 \\
        $gen^p_{mean}$ & 0.27 & 0.10 & 0.08 & \textbf{125.29} & -0.07 \\
        $line^{ex,p}_{max}$ & 0.51 & 0.21 & 0.60 & 0.58 & \textbf{1.33} \\
        $line^{ex,p}_{mean}$ & 19.16 & 12.24 & 22.07 & 19.56 & \textbf{106.85} \\
        $line^{or,p}_{max}$ & 0.51 & 0.20 & 0.61 & 0.58 & \textbf{1.33} \\
        $line^{or,p}_{mean}$ & 18.96 & 12.27 & 22.44 & 19.10 & \textbf{108.58} \\
        \bottomrule
        \end{tabular}

    \label{tab:cluster_res}
\end{table}
\begin{figure}[h]
    \centering
    \includegraphics[trim={2.0cm 3.8cm 1.0cm 2.8cm},clip, width=0.8\linewidth]{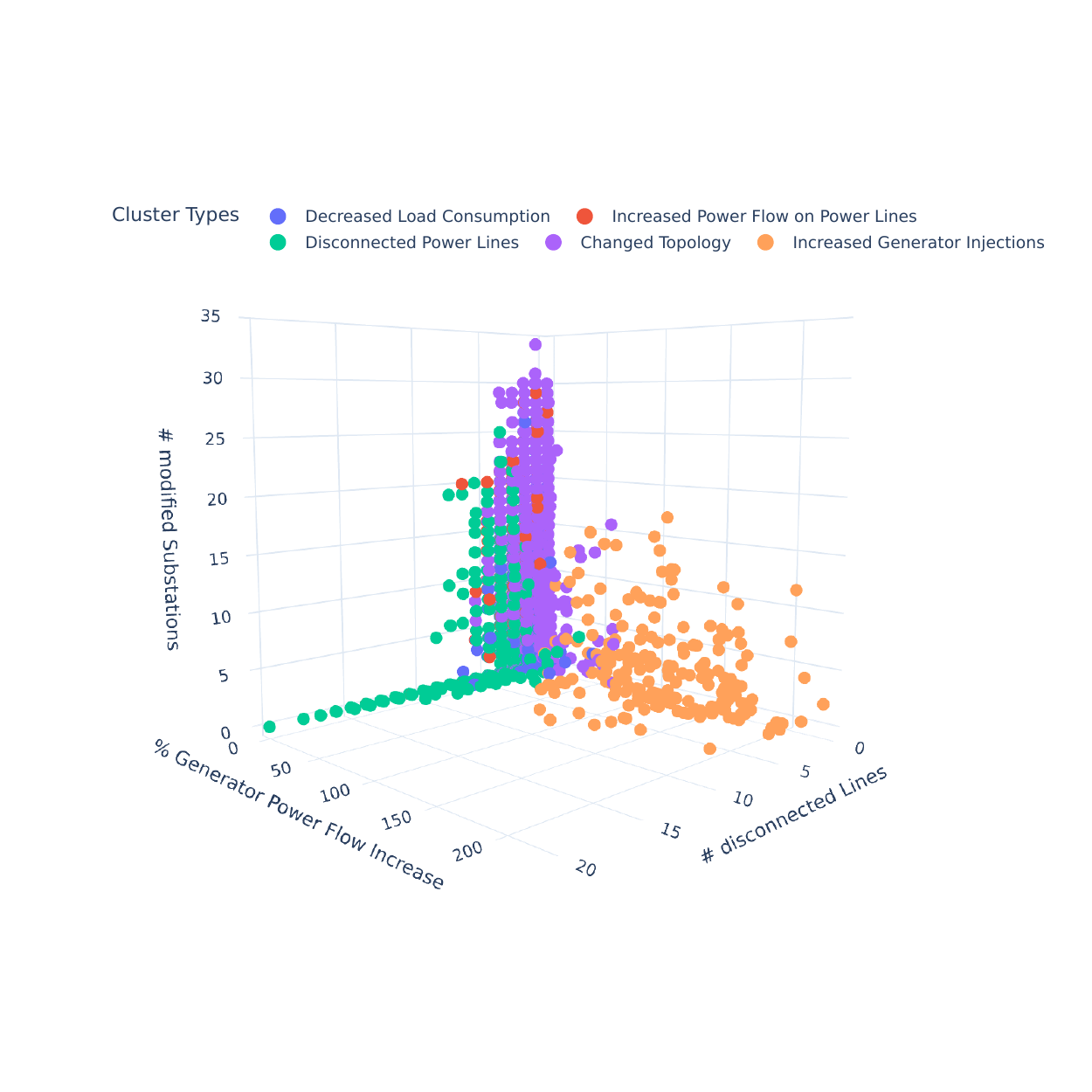}
    \caption{\small 3D visualization of the five clusters, where each point represents a failure of the agents. The axis in the plot are the $gen^p_{mean}$ on the y-axis, $\#lines_{dis}$ on the x-axis and $\#sub_{changed}$ as z-axis. The points are colored according to their respective clusters. }
    \label{fig:cluster}
\end{figure}
Interestingly, we can see in Figure \ref{fig:cluster} that the clusters \textit{Increased Generator Injections} and \textit{Disconnected Power Lines} have distinct observations that can clearly be separated from the other clusters. In contrast, there seems to be multiple clusters with an increased $\#sub_{changed}$. With respect to Table \ref{tab:cluster_res}, several noteworthy observations can be made. First, there appears to be a significant increase in generator output in the fourth cluster, with a factor of up to 125 times the $gen^p_{mean}$. Given that the relative change to $t=0$ in active power is compared, it can be explained that some generators only produce a minor amount at $t=0$ due to the fact that all scenarios start at nighttime. 
Secondly, it can be observed that the only instance of a decrease in the mean values can be found in the second cluster with $load^p_{mean}$. Given Kirchhoff's current law, i.e., the aggregated loads always have correspond to the aggregate generation in the whole grid, the lower $load^p_{mean}$ can only show localized imbalances in parts of the grid.
\begin{table}[!h]
    \caption{\small Quantity and Frequency of failures in the clusters for the three agents. We denote the number of occurrences per cluster and the percentage within brackets. We highlight for each agent the most frequent cluster. Note that the amount of failure data across the agents differs.}
    \centering
    \small
    \begin{tabular}{ccccc}
    \toprule
        & \gls{d_n} &\gls{senior} &\gls{topo} & Total \\
    \midrule 
        \makecell{Changed \\Topology}   &  1573 (10\%)  & \textbf{7191 (61\%) } & \textbf{6676 (58\%)} & 15440 (39\%) \\
        \makecell{Decreased Load\\Consumption} & 4345 (26\%)  & 2239 (19\%)& 2508 (22\%) &9092 (23\%) \\
        \makecell{Disconnected\\Power Lines} & \textbf{9225 (56\%)} & 1360 (12\%)& 1424 (12\%)& 12009 (30\%)  \\
        \makecell{Increased Generator\\Injections} & 178 (1\%)  &121 (1\%)&106 (1\%) &405(1\%)\\
        \makecell{Increased Power \\Flow on Power Lines} & 1153 (7\%)  &  784 (7\%)&752  (7\%) &2689 (7\%)\\
    \midrule 
        Total \# Obs & 16.474 &11.695 & 11.466 & 39.635 \\
    \bottomrule
    \end{tabular}
    \label{tab:cluster_agent}
\end{table}
\paragraph{Cluster distribution across Agents:} Following the general description of the clusters, we are interested in the distribution across the agents, as seen in Table \ref{tab:cluster_agent}. First, we see that \textit{Changed Topology} and \textit{Disconnected Power Lines} are quite large containing 39\% and 30\% of the data, while the \textit{Increased Generator Injections} only has a total of 405 samples. With the $\chi^2$ independence test ($\alpha=0.05$), we tested the $H_0$-hypothesis that the clusters and agents are independent from each other. We could clearly reject the hypothesis with a $p-value$ of $0.0$, which is a very strong indication that there is an association between clusters and agents. 
Comparing the clusters with the agents, we can see that a large portion of the \gls{d_n} failures are due to the disconnection of power lines (56\%) as well as the reduced load consumption (26\%). In contrast, the \gls{senior} and \gls{topo} have most of their failures in the topology cluster. Here, the \gls{topo} has with 58\% fewer topology failures, compared to the \gls{senior}'s 61\%. Instead, the \gls{topo} has 3\% more load consumption errors than the \gls{senior}.
\paragraph{Cluster Survival Time:} Besides the distribution across the agents, we are also interested in the survival time. In Figure \ref{fig:survival} we visualize the number of steps until failure with $t_{survived}$ and observe a clear difference between clusters. The \textit{Increased Generator Injections} cluster fails incredibly fast, with a median survival time of only half a day (one day equals 288 steps). The median of the \textit{Decreased Load Consumption} cluster is a little better. As contrast, we see that the \textit{Changed Topology} cluster has the longest survival time with a median value of 496 steps, which might be related to the capabilities of the \gls{topo} and \gls{senior}. For the two remaining cluster, we have a relatively similar distribution, which is understandable since both focus on line results. 
\subsection{Forecasting Results}\label{ssec:for_res}
\paragraph{Quantitative forecasting results:} In Table \ref{tab:forecasting_res}, we present the performance of all evaluated models with regard to the defined metrics of Section \ref{ssec:metric}. \gls{lgbm} outperformed the other models across all metrics, achieving an accuracy of $82\%$ and a high binary accuracy for predicting survival vs. failure of $87\%$. Interestingly, the performance remains consistent across both test and \gls{ood} data, with binary accuracy also holding steady at 87\%, indicating no data leakage and robust generalization to novel scenarios. 
In second place, the other gradient boosting approach \gls{xg} achieved a relatively good performance, however, it was not able to reach that of \gls{lgbm}. 

Compared to the gradient boosting approaches, the neural network-based methods \gls{cem} and \gls{gandalf} perform very similarly and achieve an accuracy approximately $3\%$ lower than \gls{lgbm}. Interestingly, both achieve a high accuracy for the binary problem of predicting survival vs. failure, outperforming \gls{xg}. It is also notable that both neural network-based methods achieve overall similar scores on both the test dataset and the \gls{ood} dataset. However, the scores are still lower than those achieved by \gls{lgbm} but comparable to \gls{xg}. Therefore, the \gls{gandalf} architecture did not achieve a higher performance than the \gls{cem} model, though it is more complex. In the last place is \gls{rf}, which performed the worst across all metrics.
\begin{table}[h]

\caption{\small Forecasting results of the different models on the hold-out test data. The last two columns show the balanced accuracy and binary accuracy results on the \gls{ood} test data.}
     \centering
     \small
    \begin{tabular}{l|c|c|c|c||c|c}
    \toprule
     & accuracy & \makecell{balanced\\ accuracy} & f1 micro & \makecell{binary \\ accuracy} & \makecell{\gls{ood} balanced \\ accuracy} & \makecell{\gls{ood} binary \\ accuracy} \\
    \midrule
    \gls{rf} & 0.73 & 0.62 & 0.73 & 0.82 & 0.61 & 0.82\\
    \gls{xg} & 0.80 & 0.73 & 0.80 &	0.83 & 0.73 & 0.83\\

    \gls{lgbm} & \textbf{0.82} & \textbf{0.76} & \textbf{0.82} & \textbf{0.87} & \textbf{0.76} & \textbf{0.87}\\
    \gls{cem} & 0.79 & 0.73 & 0.79 & 0.84 & 0.73 & 0.84 \\
    \gls{gandalf} & 0.79 & 0.72 & 0.79 & 0.84 & 0.72 & 0.83 \\
    \bottomrule
    \end{tabular}

\label{tab:forecasting_res}
\end{table}
With regards to the imbalanced nature of the data, it is important to also look at the balanced accuracy and the f1 scores. As expected, we observe a decrease in balanced accuracy compared to the regular accuracy score. All models except \gls{rf} exhibit a deviation of approximately $6\%$. 
\begin{figure}[!b]
    \centering
    \includegraphics[trim={1.0cm 1.0cm 0.0cm 1.5cm},clip, width=0.9\linewidth]{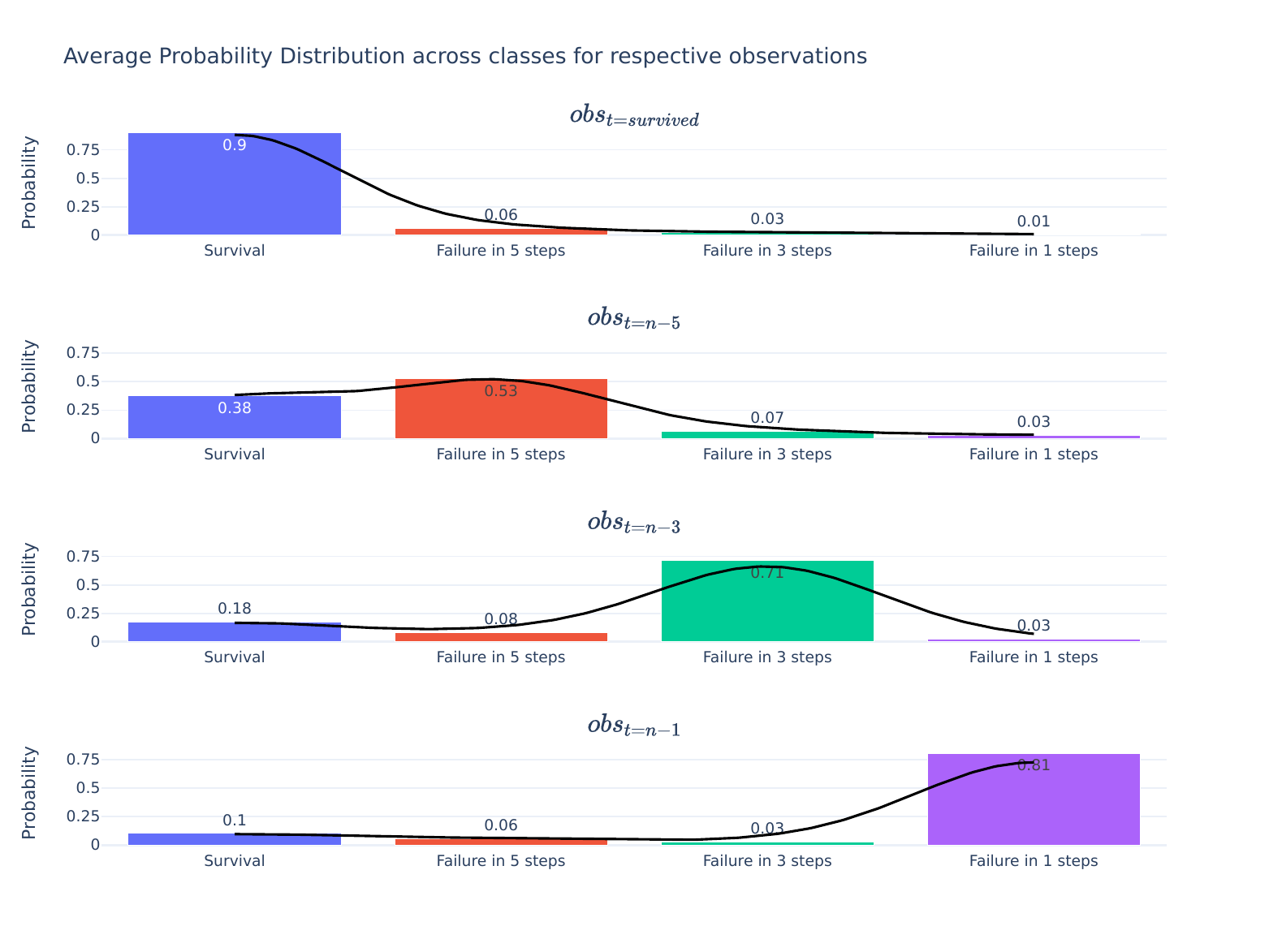}
    \caption{\small Average probability distribution of \gls{lgbm} for the ground truth of $obs_{t=survived},obs_{t=n-5},obs_{t=n-3}$ and $obs_{t=n-1}$. The probability output is averaged for all observations. Black lines visualizes the kernel density estimation across the classes.}
    \label{fig:prob_dist}
\end{figure}
\paragraph{Qualitative Forecasting result:} We take a closer look at the probability distribution of the best-performing model \gls{lgbm}. Figure \ref{fig:prob_dist} shows the averaged probability output of the model against the ground truth of all four classes. As we can see, the model selects the correct classification on average and is especially sure in the cases of survival, i.e., $obs_{survived}$ with $90\%$, and imminent failure, i.e., $obs_{t=n-1}$ with $81\%$. In the case of $obs_{t=n-5}$ we can see uncertainty in the model, since the model selects the surviving $obs_{t=survived}$ in $38\%$ of the cases.
This shows that it is difficult to distinguish $obs_{t=n-5}$ from a surviving observation, as it takes only a sudden change, such as an adversarial attack, to destabilize the grid very quickly. This can be observed for later timesteps as well, although to a lesser degree. 
\paragraph{Feature Importance}: As a last result, we examine the feature importance of the \gls{lgbm} model. For this we use the gain metric, which represents the improvement in accuracy brought by a feature to the branch. By evaluating the gain, we identify which features contribute the most to the model's predictions, allowing for a better understanding of the underlying patterns and relationships within the data. 
\begin{figure}[!h]
    \centering
    \includegraphics[trim={0.0cm 1.0cm 0.5cm 0.5cm},clip, width=1.0\linewidth]{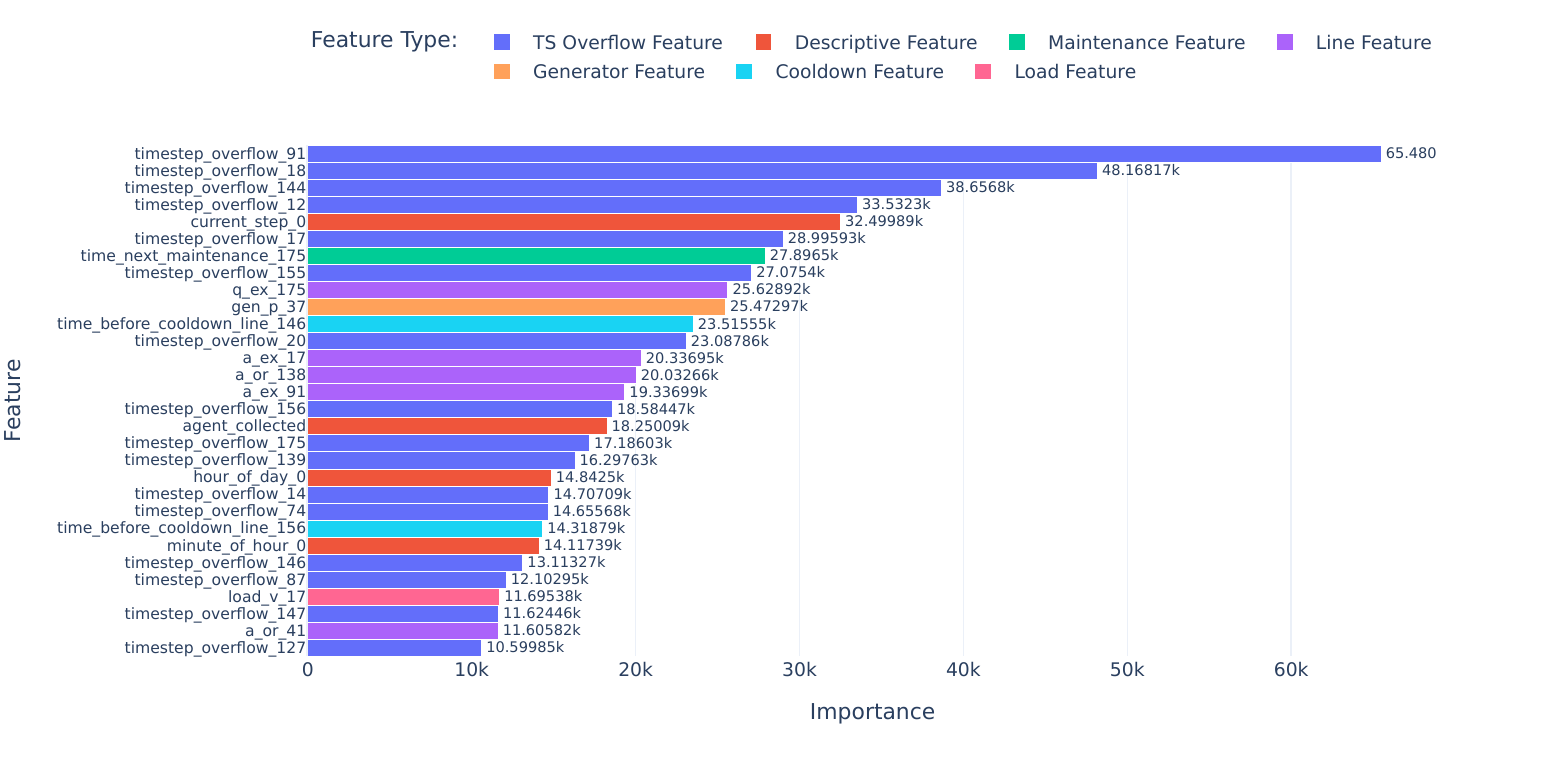}
    \caption{\small Feature importance of the 30 most important features according to the \gls{lgbm} model. The color indicates the type of variable.}
    \label{fig:feature_importance}
\end{figure}
With respect to the features of the \gls{lgbm}, we look at the top 30 features, visualized in Figure \ref{fig:feature_importance}. We group the features based on their type and distinguish among descriptive, generator, load, and line features (power flow, cooldown, $ts_{overflow}$, maintenance). 
Beginning with the interpretation of these results, we observe that $16$ of the most important features are of type $ts_{overflow}$ which denotes the number of time steps since the power line has $\rho_{l}\geq 1.0$. This is not surprising, since overloaded lines are a strong indicator for an unstable grid. However, we also have three important descriptive features ($current\_step$, $minute\_of\_hour$, and $hour\_of\_day$). These suggest that the likelihood of power grid failures is influenced by temporal features, reflecting patterns that may correlate with daily operational cycles or varying demand. Another important descriptive feature is the agent type in 17th place, since there is a clear difference in behavior between the \gls{topo}, \gls{senior} and the \gls{d_n} in terms of survival. It is plausible that some grid states would lead to a failure without remedial action, but an agent applying topological fixes could prevent this. 
Regarding the loads and generators, we can see that the first generator feature appears at position 10, while the first load feature appears only at position 27. This is quite interesting, considering that the decreased load cluster is relatively frequent. It is possible that the grid's stability is more dependent on the fluctuating nature of largely weather-dependent renewable generators than on loads. 
Examining the overall picture, it is unsurprising that the feature importance is dominated by line features. Moreover, many features correspond to the same lines. For instance, three distinct features corresponding to line 175 appear among the top 30 importance values alone. 
To simplify the interpretation of feature importance and to better understand the broader impacts of different types of grid components, we aggregated the feature importance values by averaging the importance values of related features. 
The results of the aggregation are visualized in Figure \ref{fig:grid_importance}, where we display the ten most important lines, generators and loads of the grid. 
\begin{figure}[!h]
    \centering
    \includegraphics[trim={0.5cm 2cm 0.5cm 0.5cm},clip, width=1.0\linewidth]{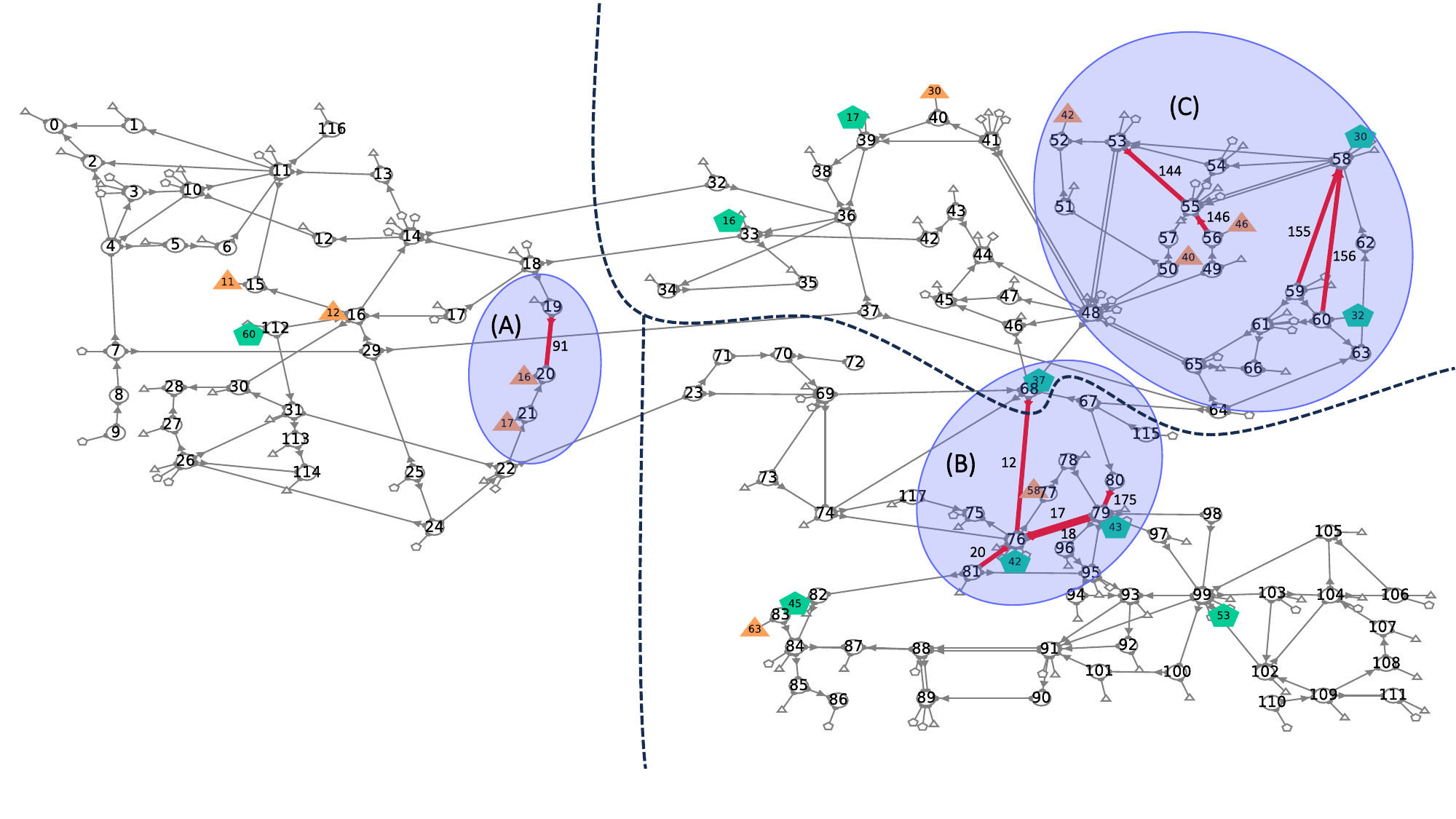}
    \caption{\small Top 10 important lines (red), generators (green), and loads (yellow) for failure prediction. The numbers on the marked lines correspond to the line ids of Figure \ref{fig:feature_importance}. Sub-grids are separated with dotted lines. 
    For better clarity, we grayed out the less important elements of the grid and highlighted 3 significant regions (A,B,C) of important grid features.
    }
    \label{fig:grid_importance}
\end{figure}

In the first region (A) we have two loads and a line of high importance. By looking at the failures we could identify the line between substations 21 and 22 as reason, as it is often targeted by the adversarial agent. This attack is a certain failure because the demand of the loads cannot be changed and the only generator that provides electricity is in substation 18. This causes an overflow in all the lines between substations 18 and 21, leading to a cascading failure in less than 3 steps. Thus, by disconnecting the line, the adversarial agent causes a failure, which to our surprise is very common with 4057 out of 39635 cases.

Region (B) highlights the connection between two sub-grids: the line between substations 68 and 76 connects the upper right sub-grid to a major substation in the bottom right sub-grid. The most important generator 37 connects to substation 68 and frequently injects up to 500\% of its starting power. Furthermore, substation 76 is connected to the central substation 79 with two lines, indicating an increased power flow and overall significance of this connection. Accordingly, these lines are deemed as very important by the prediction model. Both substations are very important since all power flow to the majority of this sub-grid have to be routed through them, hence they serve as the entry point to this part of the grid. Additionally, two other paths (80-67-64 or 80-67-68) from substation 79 to the upper right sub-grid have to traverse substation 80, hence the connection between 79 and 80 is also considered very important by the model. Overall, region (B) can be characterized as a high line power flow region, having 3 important generators (37, 42 and 43) in the vicinity, connecting two sub-grids, and being the only entry point to the bottom right area of the grid. 

Region (C) marks a section of the grid where frequent line attacks by the adversarial agent occur. In fact, 10 out of 23 possible line attacks happen in this sub-grid with 5 line attacks in region (C) alone. Both the lines connecting substations 58 to 62 and 62 to 63 are part of the attacked lines, leading to a higher load on the other lines connecting the group of substations ($59,60,61,63,64,65,66$) in the lower area of this sub-grid. Namely, these are the lines connecting substation 58 to 59 and 60, respectively, as correctly identified by the importance of the model. Moreover, three out of the ten most important loads are present in this region, connected at substations 50, 52, and 56, respectively. 
In order to supply these loads, a high load and importance at the lines connecting these loads to bigger substations is to be expected. This is exacerbated by the fact that the lines connecting substation 48 to 50 and substation 48 to 53 can be attacked, leaving less possible paths to these loads available.

\section{Discussion}\label{sec:discussion}
Looking at the results, we can see that the clustering was quite successful. 
We were able to identify at least two of the five clusters that are highly correlated with the agents, confirming our expectations. Moreover, both advanced agents were able to significantly reduce the number of failures due to line disconnection. Since the \gls{topo} has slightly less errors due to topology changes, we can also confirm \cite{lehna2024hugo}. 
In terms of the remaining clusters, the results are not as clear. For the line and generator clusters, we have very similar cluster sizes across all agents. For the load cluster, we could at least observe a slight reduction for the advanced agents. Overall, the results indicate that there is still a lot of optimization potential, requiring additional redispatching optimization. With the feature importance, we were also able to detect the design error in region (A) of Fig. \ref{fig:grid_importance}, showing a possible requirement to incorporate load shedding in future action spaces of Grid2Op. Agents should focus on preventing the disconnection of this line due to overflows at all cost by performing suitable remedial actions in a timely manner. 
With respect to our predictive models, we were able to achieve an accuracy of 82\%, showing that model correctly detect possible failures in the gird in most cases. More importantly, the binary accuracy of 87\% showed, that we are able to correctly detect whether the agent is struggling in the grid. This makes the real-world applicability more interesting, as we can alert the operator in time, with $obs_{t=n-5}$ corresponding to 25min before failure. 
As future work, improving the performance of the failure prediction could be achieved by utilizing \glspl{gnn} to capture the inherent graph structure present in the power grid topology. Further, a continuous analysis should be conducted that includes redispatching components to target the reduced load cluster by actively adjusting redispatch and curtailment. Furthermore, it might be interesting to focus on the specific regions of the grid that received high feature importance, as they seem to be responsible for the performance of the agents. This could be achieved by specifically training the \gls{drl} agent on these regions, or even a hierarchical approach. Including topological actions that specifically target these regions into the action space might bring a boost in performance as well. Finally, the prediction of failures has a potential in guiding decisions on when a DRL agent should take remedial actions based on them, rather than fixed thresholds of maximum line capacity $\rho_{max,t}$.

\section{Conclusion}\label{sec:conclusion}
In this paper, we provide a detailed failure analysis on the WCCI \gls{l2rpn} environment, which has not been done to this extent yet. We examined the failures in the grid of three different agents across ten seeds, resulting in 40k data points. Our analysis is twofold, with a detailed cluster analysis and a prediction of the failures in advance. For the cluster analysis, we were able to identify five specific clusters that showed distinct causes for the failures. The failures exhibited varying survival times and could be attributed to some extend to specific agents. As a second part, we propose a multi-class forecasting approach to detect the failures ahead of time. On the data we tested five different model types and found the \gls{lgbm} to be the most suitable model for the prediction. The feature importance afterwards revealed critical regions of the grid that could be specifically targeted in future work. 
\section*{Acknowledgement}
This work was supported by (i) the research group Reinforcement Learning for Cognitive Energy Systems (RL4CES) founded by the German Federal Ministry of Education and Research (01|S22063B), (ii) Graph Neural Networks for Grid Control (GNN4GC) founded by the Federal Ministry for Economic Affairs and Climate Action Germany(03EI6117A) and (iii) AI4REALNET founded by the European Union’s Horizon Europe Research and Innovation program under the Grant Agreement No 101119527. Views and opinions expressed are however those of the author(s) only and do not necessarily reflect those of the European Union. Neither the European Union nor the granting authority can be held responsible for them.

\section*{Competing Interest}
The authors have no competing interests to declare that are
relevant to the content of this article.

%
%
%
\bibliographystyle{splncs04}
\bibliography{hugo_bib}

\begin{thebibliography}{10}
\providecommand{\url}[1]{\texttt{#1}}
\providecommand{\urlprefix}{URL }
\providecommand{\doi}[1]{https://doi.org/#1}

\bibitem{abdi2010principal}
Abdi, H., Williams, L.J.: Principal component analysis. Wiley interdisciplinary reviews: computational statistics  (2010)

\bibitem{optuna}
Akiba, T., Sano, S., et~al.: Optuna: A next-generation hyperparameter optimization framework. In: Proceedings of the 25th {ACM} {SIGKDD} International Conference on Knowledge Discovery and Data Mining (2019)

\bibitem{amdouni2023grid2onto}
Amdouni, E., Khouadjia, M., et~al.: Grid2onto: An application ontology for knowledge capitalisation to assist power grid operators. In: International Conference On Formal Ontology in Information Systems-Ontology showcases and Demos (2023)

\bibitem{bacher1986network}
Bacher, R., Glavitsch, H.: Network topology optimization with security constraints. IEEE Transactions on Power Systems  (1986)

\bibitem{chauhan2022powrl}
Chauhan, A., Baranwal, M., et~al.: Powrl: A reinforcement learning framework for robust management of power networks. arXiv preprint arXiv:2212.02397  (2022)

\bibitem{chen2016fault}
Chen, K., Huang, C., et~al.: Fault detection, classification and location for transmission lines and distribution systems: a review on the methods. High voltage  (2016)

\bibitem{xgboost}
Chen, T., Guestrin, C.: Xgboost: A scalable tree boosting system. In: Proceedings of the 22nd acm sigkdd international conference on knowledge discovery and data mining (2016)

\bibitem{dashti2021survey}
Dashti, R., Daisy, M., et~al.: A survey of fault prediction and location methods in electrical energy distribution networks. Measurement  (2021)

\bibitem{ding2004k}
Ding, C., He, X.: K-means clustering via principal component analysis. In: Proceedings of the twenty-first international conference on Machine learning (2004)

\bibitem{dorfer2022power}
Dorfer, M., Fuxj{\"a}ger, A.R., et~al.: Power grid congestion management via topology optimization with alphazero. arXiv preprint arXiv:2211.05612  (2022)

\bibitem{binbinchen}
{EI Innovation Lab, Huawei Cloud, Huawei Technologies}: {NeurIPS Competition 2020: Learning to Run a Power Network (L2RPN) - Robustness Track}. \url{https://github.com/AsprinChina/L2RPN\_NIPS\_2020\_a\_PPO\_Solution} (2020), accessed 22-Jan-2023 on Github

\bibitem{grlsurvey2024}
Hassouna, M., Holzhüter, C., Lytaev, P., Thomas, J., Sick, B., Scholz, C.: Graph reinforcement learning for power grids: A comprehensive survey (2024), \url{https://arxiv.org/abs/2407.04522}

\bibitem{rf}
Ho, T.K.: Random decision forests. In: Proceedings of 3rd international conference on document analysis and recognition. IEEE (1995)

\bibitem{pytorchtabular}
Joseph, M.: Pytorch tabular: A framework for deep learning with tabular data (2021)

\bibitem{gandalf}
Joseph, M., Raj, H.: Gandalf: Gated adaptive network for deep automated learning of features (2024)

\bibitem{lgbm}
Ke, G., Meng, Q., et~al.: Lightgbm: A highly efficient gradient boosting decision tree. Advances in neural information processing systems  (2017)

\bibitem{kelly2020reinforcement}
Kelly, A., O'Sullivan, A., et~al.: Reinforcement learning for electricity network operation. arXiv preprint arXiv:2003.07339  (2020)

\bibitem{kodinariya2013review}
Kodinariya, T.M., Makwana, P.R., et~al.: Review on determining number of cluster in k-means clustering. International Journal  (2013)

\bibitem{lehna2024hugo}
Lehna, M., Holzh{\"u}ter, C., et~al.: Hugo--highlighting unseen grid options: Combining deep reinforcement learning with a heuristic target topology approach. arXiv preprint arXiv:2405.00629  (2024)

\bibitem{lehna2023managing}
Lehna, M., Viebahn, J., et~al.: Managing power grids through topology actions: A comparative study between advanced rule-based and reinforcement learning agents. Energy and AI  (2023)

\bibitem{macqueen1967some}
MacQueen, J., et~al.: Some methods for classification and analysis of multivariate observations. In: Proceedings of the fifth Berkeley symposium on mathematical statistics and probability (1967)

\bibitem{marot2020learning}
Marot, A., Donnot, B., et~al.: Learning to run a power network challenge for training topology controllers. Electric Power Systems Research  (2020a)

\bibitem{marot2021learning}
Marot, A., Donnot, B., et~al.: Learning to run a power network challenge: a retrospective analysis. In: NeurIPS 2020 Competition and Demonstration Track. PMLR (2021)

\bibitem{marot2022learning}
Marot, A., Donnot, B., et~al.: Learning to run a power network with trust. Electric Power Systems Research  (2022)

\bibitem{marot2020l2rpn}
Marot, A., Guyon, I., et~al.: L2rpn: Learning to run a power network in a sustainable world neurips2020 challenge design (2020b)

\bibitem{mu2022cascading}
Mu, Z., Xu, P., et~al.: Cascading fault early warning and location method of transmission networks based on wide area time-series power system state. IEEE Journal of Radio Frequency Identification  (2022)

\bibitem{serre2022reinforcement}
Serr{\'e}, G., Boguslawski, E., et~al.: Reinforcement learning for energies of the future and carbon neutrality: a challenge design. arXiv preprint arXiv:2207.10330  (2022)

\bibitem{9494879}
Subramanian, M., Viebahn, J., et~al.: Exploring grid topology reconfiguration using a simple deep reinforcement learning approach. In: 2021 IEEE Madrid PowerTech (2021)

\bibitem{viebahn2022potential}
Viebahn, J., Naglic, M., et~al.: Potential and challenges of {AI}-powered decision support for short-term system operations. CIGRE 2022 Paris Session  (2022)

\bibitem{tpe}
Watanabe, S.: Tree-structured parzen estimator: Understanding its algorithm components and their roles for better empirical performance (2023)

\bibitem{zhou2021action}
Zhou, B., Zeng, H., et~al.: Action set based policy optimization for safe power grid management. In: Machine Learning and Knowledge Discovery in Databases. Applied Data Science Track: European Conference, ECML PKDD 2021 (2021)

\end{thebibliography}
\newpage


\section{Appendix}
\subsection{Cluster Variables}
\label{app:cluster_variables}
\begin{figure}[!h]
    \centering
    \includegraphics[trim={0.0cm 1.0cm 2.0cm 0.0cm},clip, width=0.7\linewidth]{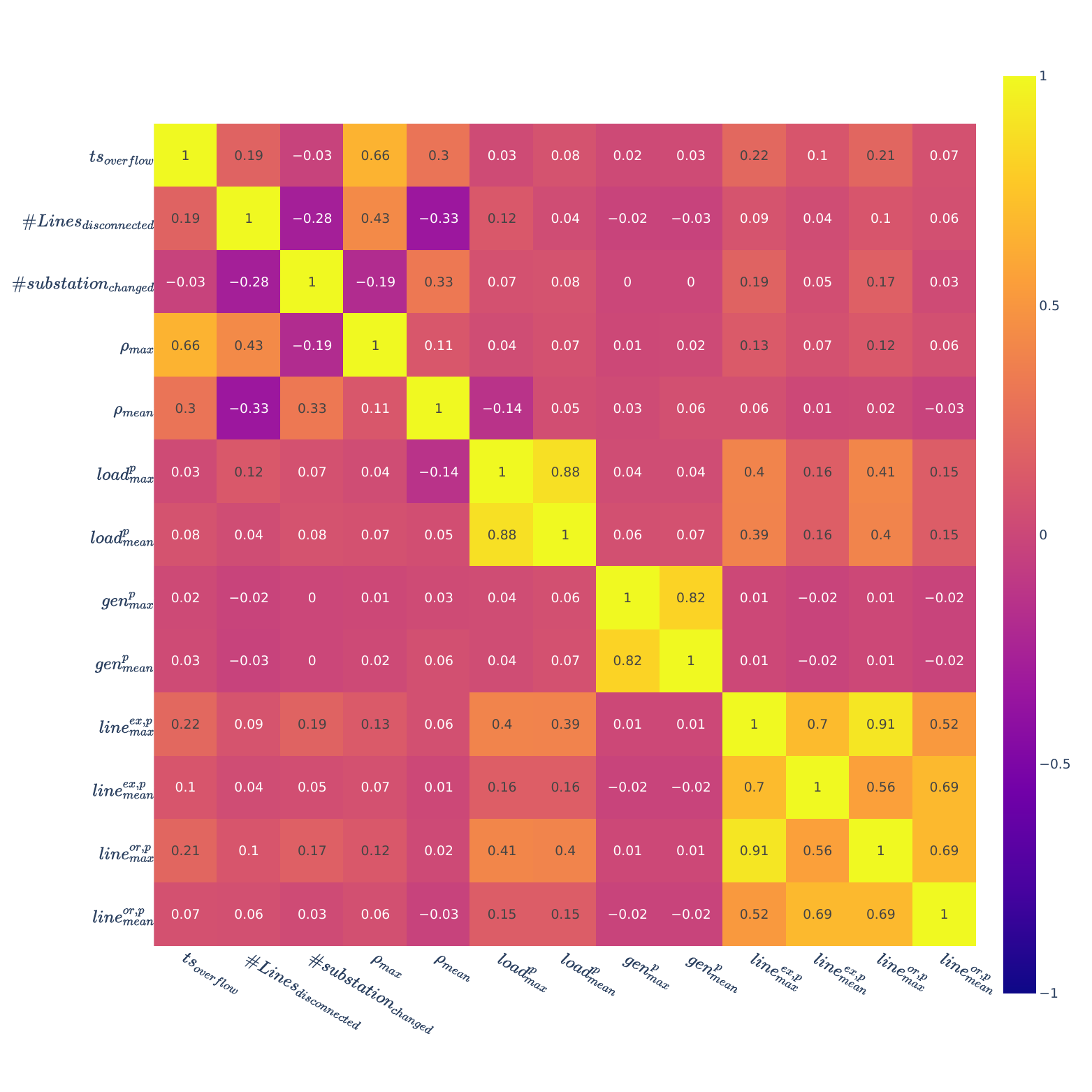}
    \caption{Correlation plot of the different cluster variables. The color range depicts the correlation values that go from -1.0 (blue) to 1.0 (yellow).}

    \label{fig:correlation}
\end{figure}

We summarize the distribution of the cluster variables in the following Table \ref{tab:cluster_variables}.
 \begin{table}[!h]
\centering
\small
\begin{tabular}{lcrrrrrrrr}
    \toprule
    Name & Explanation & mean & std & min & 25\% & 50\% & 75\% & max \\
    \midrule
    $ts_{overflow}$ & \makecell{Cumulative \\overflow time}& 8.25 & 6.78 & 0.0 & 3.0 & 6.0 & 12.0 & 71.0 \\
    $\#lines_{dis}$  & \makecell{Number of\\disconnected lines} & 2.02 & 2.27 & 0.0 & 0.0 & 1.0 & 3.0 & 20.0 \\
   $\#sub_{changed}$ & \makecell{Number of changed\\Substations}  & 3.10 & 4.63 & 0.0 & 0.0 & 1.0 & 4.0 & 34.0 \\
    $\rho_{max}$ & Max line capacity & 1.19 & 0.28 & 0.61 & 1.03 & 1.11 & 1.33 & 2.0 \\
    $\rho_{mean}$ & Mean line capacity & 0.38 & 0.06 & 0.19 & 0.33 & 0.38 & 0.42 & 0.63 \\
    $load^p_{max}$ & \makecell{Max active load\\value of consumption}  & 0.28 & 0.14 & -0.14 & 0.22 & 0.30 & 0.37 & 0.88 \\
    $load^p_{mean}$ & \makecell{Mean active load\\value of consumption}  & 0.08 & 0.10 & -0.26 & 0.04 & 0.11 & 0.15 & 0.32 \\
  $gen^p_{max}$ &  \makecell{Max active production\\value of genreators}  & 10.03 & 70.03 & -0.48 & 0.79 & 1.67 & 3.27 & 1573.0 \\
    $gen^p_{mean}$ & \makecell{Mean active production \\value of genreators} & 1.43 & 13.40 & -0.98 & -0.24 & -0.0 & 0.30 & 224.02 \\
   $line^{ex,p}_{max}$ &\makecell{Max active power\\flow at line extremity} & 0.52 & 0.39 & -0.22 & 0.24 & 0.46 & 0.73 & 4.60 \\
   $line^{ex,p}_{mean}$ & \makecell{Mean active power\\ flow at line extremity } & 24.41 & 32.13 & 0.24 & 8.60 & 15.89 & 27.25 & 696.11 \\
 $line^{or,p}_{max}$ & \makecell{Max active power\\flow at line origin}   & 0.53 & 0.39 & -0.22 & 0.24 & 0.46 & 0.74 & 5.64 \\
  $line^{or,p}_{mean}$ & \makecell{Mean active power\\flow at line origin}   & 24.56 & 33.86 & 0.23 & 8.57 & 15.84 & 27.15 & 730.49 \\
  \midrule
  $t_{survived}$ & Survival Times &539.87 & 470.42 & 2.0 & 147.0 & 407.0 & 804.0 & 2016.0 \\
    \bottomrule
    \end{tabular}
\caption{\small Variables for clustering. Note that $t_{survived}$ is not included in the clustering and is just reported to showcase the distribution.}
\label{tab:cluster_variables}
\end{table}
\paragraph{Correlation of cluster variables:} When looking at the correlation plot in Figure \ref{fig:correlation}, there are already three interesting things to note. First, we see a high correlation in the power flow values between the maximum and average components, which is relatively self-explanatory. Second, we can see that the load values are somewhat correlated to $line^{ex,p}_{max},line^{or,p}_{max}$. Interestingly, the generator values are not correlated with the line power flow in any way.  Third, we see a negative correlation between $\#sub_{changed}$ and $\#lines_{dis}$, which can be explained by the fact that both the \gls{senior} and \gls{topo} agents have components that automatically reconnect lines. However, $\#lines_{dis}$ is also negatively correlated with $\rho_{max}$, which is counter-intuitive. A possible explanation could be the Grid2Op rule that disconnects a line after three consecutive time steps of overload. When the line is disconnected, the highest $\rho_{max}$ is no longer available, explaining the negative correlation. This explanation would primarily hold for consecutive observations, so it is quite interesting that it seems to work across multiple scenarios.

\subsection{Inertia and Silhouette plot}
\begin{figure} [!h]
    \centering
    \includegraphics[trim={0.0cm 1.0cm 0.0cm 2.0cm},clip, width=1.0\linewidth]{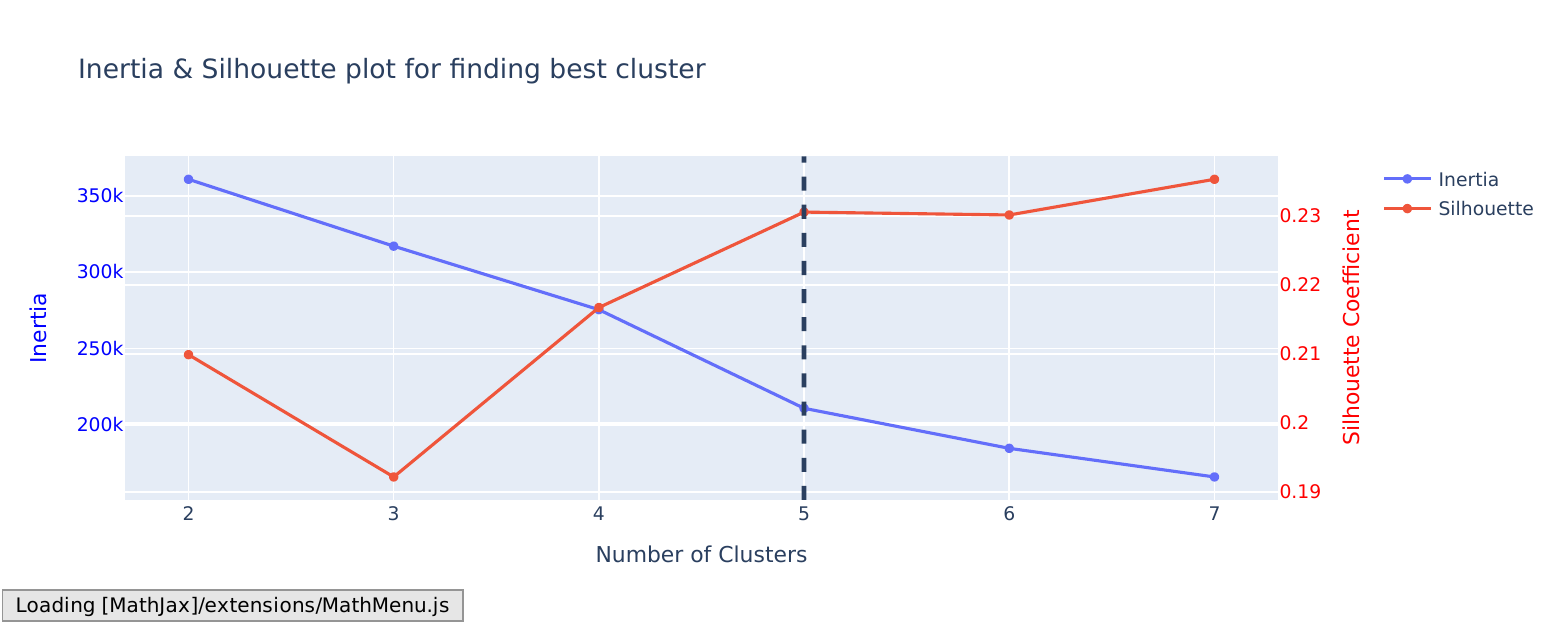}
    \caption{\small Inertia and Silhouette plot of different cluster sizes. As one can see, the inertia has a "elbow" at a cluster size of five. Similar, the silhouette shows one of its highest values at a size of five as well. Thus, we select a cluster size of five.}
    \label{fig:intertia}
\end{figure}

The plot below (Fig. \ref{fig:intertia}) illustrates the inertia values for different cluster sizes in our k-means clustering analysis, showcasing the selection process for determining the optimal number of clusters based on the elbow method and the corresponding decrease in inertia.

\label{ssec:inertia}


\subsection{Additional Information}
    
\label{app:error_statement}
\paragraph{Error Statement}
We collected the error statement of every failed scenario to see, whether we could detect a pattern. The following three error statements were outputted by the underlying Grid2Op solver \verb|lightsim2grid|:
\begin{enumerate}
    \item \texttt{Grid2OpException Divergingpower flow "Divergence of DC power flow (non connected grid) at the initialization of AC power flow. Detailed error: ErrorType.SolverFactor"}
    \item \texttt{Grid2OpException Divergingpower flow "Divergence of AC power flow. Detailed error: ErrorType.TooManyIterations"}
    \item \texttt{Grid2OpException Divergingpower flow "Divergence of DC power flow (non connected grid) at the initialization of AC power flow. Detailed error: ErrorType.SolverSolve"}
\end{enumerate}
We collected all error messages and mapped them to the clusters of Section \ref{ssec:cluster}. The results can be found in Table \ref{tab:error_cluster_mapped}. While there is some difference in the percentage change, we can not find a distinct relation between cluster and error message.

\begin{table}[!h]
    \centering
    \small
    \begin{tabular}{lrrrrr}
        \toprule
        \small
        \makecell{Error \\Type} &  \makecell{Changed \\Topology}    & \makecell{Decreased Load\\Consumption} & \makecell{Disconnected\\Power Lines} & \makecell{Increased Generator\\Injections} & \makecell{Increased power flow\\on Power Lines}  \\
        \midrule
        1 & 61.2\% & 63.97\% & 69.27\% & 68.15\% & 65.01\% \\
        2 & 27.47\% & 27.03\% & 24.71\% & 26.42\% & 26.18\% \\
        3 & 11.32\% & 9\% & 6.01 & 5.43\% & 8.81\% \\
        \bottomrule
\end{tabular}
    \caption{In this Table, we map the three error types with the different clusters. For each cluster we aggregate the error types to percentage values for better viability.}
    \label{tab:error_cluster_mapped}

\end{table}
\paragraph{Specific Descriptive Scenarios}
We further accounted for three descriptive variables for the scenarios: 
\begin{itemize}
    \item $Attack_{93}$: This variables indicates, whether the Line 93 is disconnected or will be disconnected in the next step 
    \item This variable indicates, whether the grid was in its original state, i.e., all substation had their buses set to one and all lines were disconnected 
    \item Lastly, we collected the percentage change of the generator 37, as this generator was directly at substation 68.
\end{itemize}

For the first two variables, we have the following occurence: 
\subsection*{Specific Descriptive Scenarios}
\begin{table}[!h]
    \centering
    \small

    \begin{tabular}{c|c|c|c|c|}
    \toprule
        & \gls{d_n} &\gls{senior} &\gls{topo} & Total \\
    \midrule 
        $Attack_{93}$ & 1367& 1312 &  1378 &4057 \\
        $Stable$ & 169 & 817&646 &1632\\
    \bottomrule
    \end{tabular}
    \caption{Specific failure cases identified in descriptive analysis.}
    \label{tab:specfic_cases}
\end{table}

In case of the generator 37, we can see the following percentage increase across the scenarios: 
\begin{figure}[!h]
    \centering
    \includegraphics[trim={0.0cm 2.0cm 0.0cm 2.0cm},clip, width=0.80\linewidth]{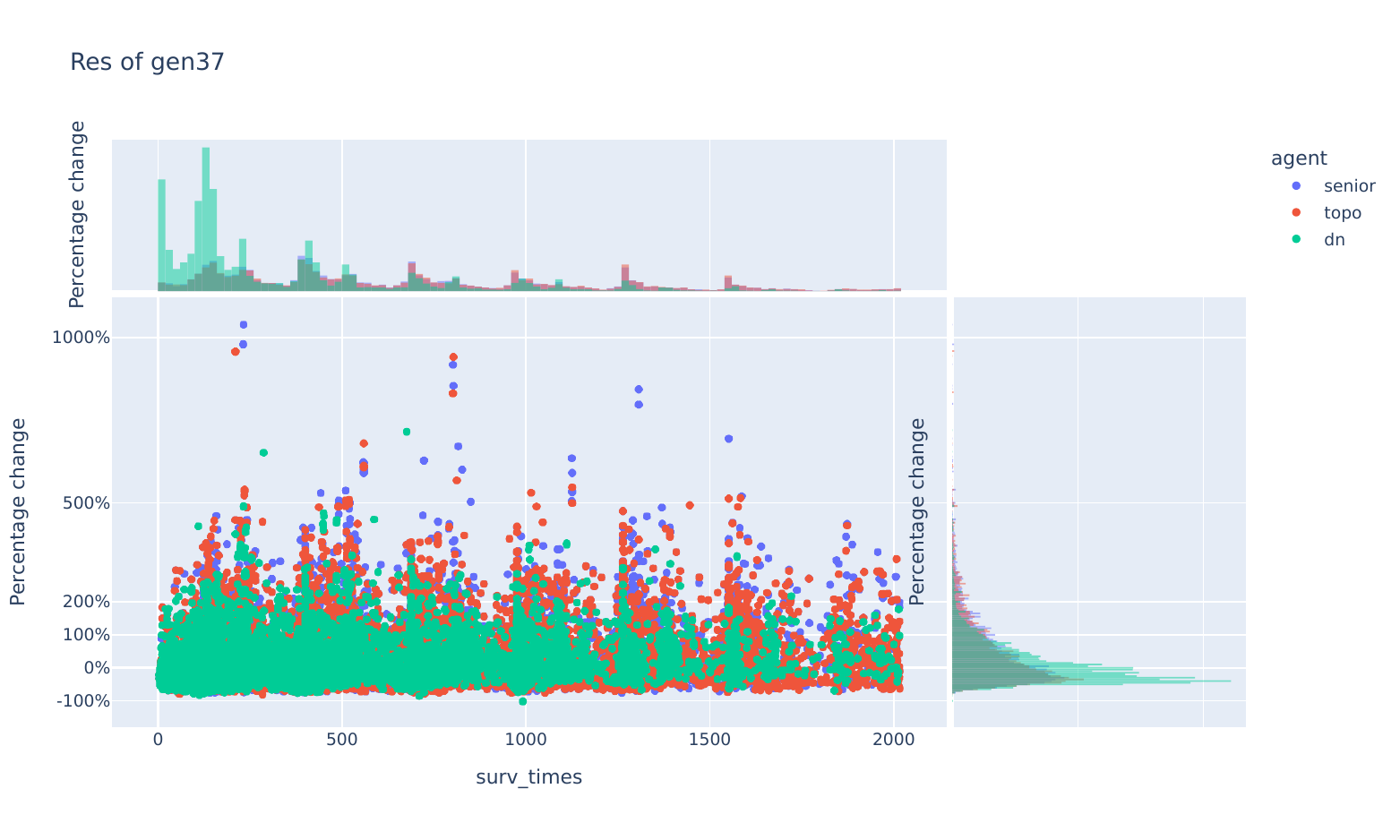}
    \caption{Visualization of the percentage change on the generator 37 in comparison to its $obs_{t=0}$.}
    \label{fig:gen37}
\end{figure}

\end{document}